\documentclass{article}

\usepackage{microtype}
\usepackage{graphicx}
\usepackage{subfigure}
\usepackage{booktabs} %
\usepackage{multirow}
\usepackage{bm}
\usepackage{hyperref}

\usepackage[accepted]{icml2024}

\usepackage{amsmath}
\usepackage{amssymb}
\usepackage{mathtools}
\usepackage{amsthm}

\usepackage[capitalize,noabbrev]{cleveref}

\theoremstyle{plain}

\theoremstyle{definition}

\theoremstyle{remark}

\usepackage[textsize=tiny]{todonotes}

\icmltitlerunning{VarteX: Enhancing Weather Forecast through Distributed Variable Representation}

\begin{document}

\twocolumn[
\icmltitle{VarteX: Enhancing Weather Forecast through Distributed Variable Representation}

\icmlsetsymbol{equal}{*}

\begin{icmlauthorlist}
\icmlauthor{Ayumu Ueyama}{student}
\icmlauthor{Kazuhiko Kawamoto}{professor}
\icmlauthor{Hiroshi Kera}{professor}
\end{icmlauthorlist}

\icmlaffiliation{student}{Graduate School of Science and Engineering, Chiba University}
\icmlaffiliation{professor}{Graduate School of Informatics, Chiba University}

\icmlcorrespondingauthor{Hiroshi Kera}{kera@chiba-u.jp}

\icmlkeywords{Machine Learning, ICML}

\vskip 0.3in
]

\printAffiliationsAndNotice{} %

\begin{abstract}
Weather forecasting is essential for various human activities. Recent data-driven models have outperformed numerical weather prediction by utilizing deep learning in forecasting performance. However, challenges remain in efficiently handling multiple meteorological variables. This study proposes a new variable aggregation
scheme and an efficient learning framework for that challenge. Experiments show that VarteX outperforms the conventional model in forecast performance, requiring significantly fewer parameters and resources. The effectiveness of learning through multiple aggregations and regional split training is demonstrated, enabling more efficient and accurate deep learning-based weather forecasting.
\end{abstract}

\section{Introduction}\label{submission}
From strategies addressing extreme weather to daily societal activities, weather forecasting plays an indispensable role in human activities \cite{bauer2015quiet}. Recently, there has been increasing interest in applying data-driven models utilizing deep learning for weather forecasting ~\cite{scher2019weather,weyn2019can, rasp2020weatherbench,weyn2021sub,keisler2022forecasting,lam2023learning}. 
With intensive training on meteorological data, such models can generate forecasts within seconds~\cite{lynch2008origins}, whereas numerical weather prediction needs to solve complex partial differential equations, leading to significantly longer forecasting time. 
Several recent studies have reported that data-driven models outperform numerical weather prediction models even in the foretasting ability~\cite{bi2023pangu, lam2023learning, chen2023fuxi}.

Many data-driven weather forecasting models~\cite{pathak2022fourcastnet, bi2023pangu,chen2023fengwu,chen2023fuxi,man2023wmae,nguyen2023climax, ni2023kunyu, nguyen2024scaling, ramavajjala2024heal} are based on Vision Transformer~(ViT;~\citet{dosovitskiy2021vit}), a powerful attention-based model in computer vision. 
This is because meteorological data closely resemble image data in their structure, having height, width, and channel dimensions.
A critical difference lies in the channel dimension. Image data only has RGB channels, which share similar information about the entire image. In contrast, meteorological data have many more channels for meteorological variables, such as temperature and humidity, with unique characteristics. The large number of meteorological variables increases the computational costs, and their diversity makes learning challenging.
ClimaX~\cite{nguyen2023climax} addressed this challenge with minimal modifications in ViT. Particularly, it equips a variable aggregation model that sums up meteorological variables into one representative variable with attention weights. Their experiments show that such an input-dependent variable aggregation leads to more successful training than the input-agnostic convolution of variables, a standard method in the image domain to summarize RGB channels.

In this study, we propose a new variable aggregation scheme and training method for efficient learning from meteorological data with ViT-based weather forecasting model. 
Our variable aggregation scheme is based on the hypothesis that the representative variable (or its $D$-dimensional embedding vector) obtained by ClimaX variable aggregation may internally contain several components because, otherwise, it is too restrictive. If so, it is better to explicitly model them as $R>1$ representative variables with $D/R$-dimensional embedding vectors. During the encoding process by our model, these embedding vectors are separately processed by Transformer encoders and then mixed by a mixing layer. The smaller embedding dimension leads to a smaller Transformer encoder, reducing the number of parameters by $1/R^2$ for each. While this reduces the model size, the memory cost at forward pass remains unchanged because the size of the attention map is determined by the number of tokens (i.e., image patches). To address this, we introduce regional split training, where a model is trained only on a cropped region at a single forward. This training decreases the final accuracy, but with a proper choice of crop size ratio $S$, the decrease is moderate, and the training time and spacial cost are reduced drastically in return. Specifically, the space complexity reduction follows $\mathcal{O}(1/S^2)$.
In experiments, we trained our model, VarteX, as well as ClimaX, on the WeatherBench dataset from scratch. The results show that VarteX forecasting accuracy is 50\% higher on average than that of ClimaX, and the gap is even larger for wind speed forecasting.
Regarding latitude-weighted root mean squared error (RMSE) and latitude-weighted anomaly correlation coefficient (ACC) for all target variables, learning effectiveness is highlighted through multiple representative variable aggregations. 
VarteX achieved these results with 55\% model size, 50\% training time, and 35\% memory usage than ClimaX.

\section{Problem setting}
Meteorological data is a time-series data of meteorological variables, such as temperature, geopotential, and wind speed, at each gird of the world. Suppose we have $H \times W$ grids and $V$ variables, then we have $ X_t \in \mathbb{R}^{H\times W \times V}$ at time step $t$. The learning-based weather forecasting aims to find a forecasting function $f_{\theta}: X_t \mapsto X_{t +\Delta t}$ with a predesignated lead time $\Delta t$ through the training of deep learning model with parameters $\theta$ in the regression task. 

An important characteristic of meteorological data is its many variables (e.g., $V= 48$). Using attention-based models such as Vision Transformer means we have to handle as many as $N = HWV$ tokens, and the attention computation grows quadratically concerning $N$ to capture the spacial and inter-variable interactions. 
As each meteorological variable has its own time- and space-dynamical characteristics, we cannot resort to a simple concatenation of them as we do on the R, G, and B variables in image data. 

The focus of our study lies in the aggregation of meteorological variables. 
Particularly, we are interested in aggregation function $\mathcal{A}: X_t \in \mathbb{R}^{H\times W \times V} \mapsto Z_t \in \mathbb{R}^{H\times W \times R}$, which reduces variables from $V$ to $R$ while achieving successful learning. ClimaX offers such an aggregation function with $R=1$. We suspect that reducing a single variable may be too aggressive and extend its idea to general $R$. 
However, as the number of $R$ increases, the number of matrices for the attention mechanism also increases, leading to a rise in computational cost. Therefore, we should also address the reduction in computational cost.

\section{Methodology}
We propose a new variable aggregation scheme and efficient training framework. As demonstrated in \cref{sec:experiments}, the former reduces the model size and significantly improves the prediction performance, and the latter realizes the training in half or even less time and memory cost. 

\subsection{Variable aggregation in ClimaX}
First, we review the cross-attention-based variable aggregation of ClimaX. 
Let $\widetilde{X} \in \mathbb{R}^{H\times W \times V \times D}$ be an input data embedded to $D$-dimensional space. 
In the following, the same operations are spatially uniformly applied, so we focus on position $(h, w) \in \{1, \ldots, H\} \times \{1, \ldots, W\}$ for notional simplicity, and re-define $\widetilde{X}$ by $\widetilde{X}_{hw} \in \mathbb{R}^{V \times D}$.
With a trainable query vector $\bm{q}\in\mathbb{R}^{D}$, the cross-attention is computed to aggregate the variables as follows.
\begin{align}
    \bm{z}^{\top} = \mathrm{softmax}\left(\frac{\bm{q}^{\top} \widetilde{K}^{\top}}{\sqrt{D}}\right)\widetilde{V} \in \mathbb{R}^{1\times D},
\end{align}
where $\widetilde{K} = \widetilde{X}W_{\mathrm{K}}$ and $\widetilde{V} = \widetilde{X}W_{\mathrm{V}}$ with trainable weights $W_{\mathrm{K}},W_{\mathrm{V}} \in \mathbb{R}^{D\times D}$ and $\mathrm{softmax}(\,\cdot\,)$ is the softmax operation. 
Namely, this cross-attention computes a weighted sum of a linear transformation of the input, $\widetilde{V} = \widetilde{X}W_{\mathrm{V}}$ with the attention weights $\bm{a}^{\top} = \mathrm{softmax}\left(\frac{\bm{q}^{\top} \widetilde{K}^{\top}}{\sqrt{D}}\right)$, thereby aggregating $V$ variables into one representative variable.

\subsection{Model architecture}

\begin{figure}[t]
\vskip 0.2in
\begin{center}
\centerline{\includegraphics[width=\columnwidth]{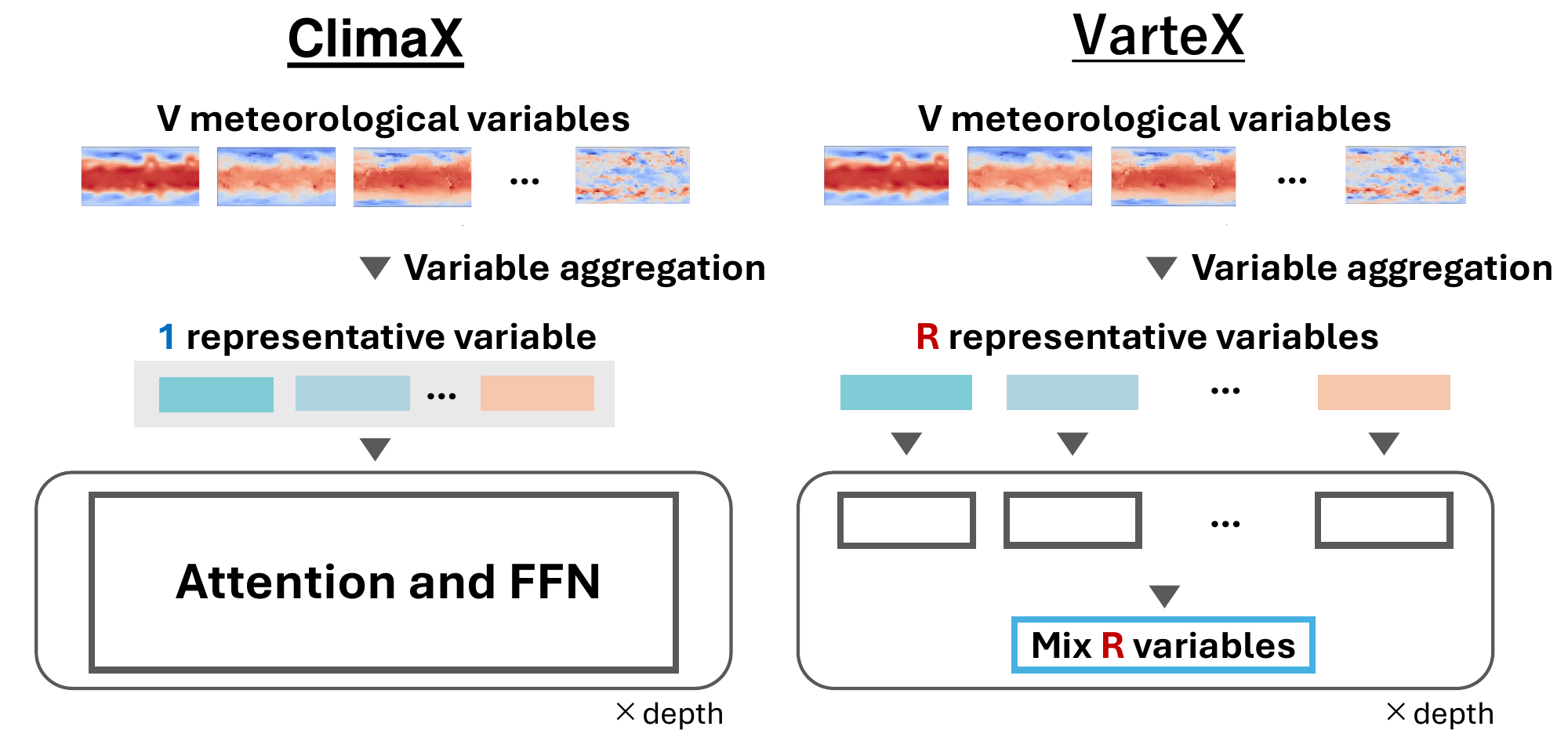}}
\caption{
Comparison of ClimaX and VarteX architectures. ClimaX aggregates V meteorological variables into a single representative variable, whereas VarteX aggregates them into R representative variables. VarteX has a layer for learning each representative variable and for learning a mixture of representative variables.}
\label{Fig:vartex}
\end{center}
\vskip -0.2in
\end{figure}

\begin{table*}[t]
    \centering
    \caption{Comparison of the proposed model (VarteX) and ClimaX trained on the ERA5 dataset from scratch (no pretraining). VarteX has $R$ representative variables, each of which is embedded into $(D/R)$-dimensional space. The only exception is $(R=4)^*$, where the embedding dimension of a representative variable is doubled to $D=2048$.}
    \label{Tab:Comparison_climax_vartex}
    \vskip 0.15in
    \begin{tabular}{llccccccccc}
        \toprule
        Lead &
        \multirow{2}{*}{Model} &
        \multicolumn{2}{c}{U10} &
        \multicolumn{2}{c}{T2m} &
        \multicolumn{2}{c}{Z500} &
        \multicolumn{2}{c}{T850} &
        \multirow{1}{*}{Parameters}  \\
        time&& ACC $\uparrow$ & RMSE $\downarrow$& ACC $\uparrow$ & RMSE $\downarrow$ & ACC $\uparrow$ & RMSE $\downarrow$ & ACC $\uparrow$ & RMSE $\downarrow$  & (M) \\ \midrule
        \multirow{4}{*}{6h}&ClimaX & 0.59 & 3.35 & 0.66 & 4.19 & 0.76 & 667.34 & 0.71 & 3.53 & 108.08  \\ 
        &VarteX~$(R=2)$ & \textbf{0.89} & \textbf{1.89}  & \textbf{0.79} & \textbf{3.24} & \textbf{0.97} & \textbf{247.33} & \textbf{0.92} & \textbf{1.90} & 59.86  \\
        &VarteX~$(R=4)$ & 0.19 & 4.54  & 0.32 & 7.43 & 0.53 & 966.76 &0.51 & 4.65 & 20.76 \\ 
        &VarteX~$(R=4)^*$ & 0.76 & 2.66  & 0.56 & 5.13 & 0.81 & 599.42 & 0.80 & 2.93& 80.47  \\ \midrule
        \multirow{4}{*}{24h}&ClimaX & 0.37 & 3.86 & 0.66 & 4.20 & 0.69 & 734.47 & 0.70 & 3.58 & 108.08   \\ 
        &VarteX~($R=2)$ & \textbf{0.64} & \textbf{3.16}  & \textbf{0.83} & \textbf{2.99} & \textbf{0.88} & \textbf{478.42} & \textbf{0.85} & \textbf{2.59}&  59.86  \\
        &VarteX~$(R=4)$ & 0.10 & 4.76  & 0.31 & 7.33 & 0.48 & 986.85 & 0.47 & 4.77&  20.76  \\
        &VarteX~$(R=4)^*$ & 0.52 & 3.52  & 0.57 & 4.99 & 0.75 & 660.40 & 0.75 & 3.25&   80.47\\
        \bottomrule
    \end{tabular}
    \vskip -0.1in
\end{table*}

While ClimaX variable aggregation reduces $V$ variables into one, we hypothesize that the representative variable (and its embedding vector) may internally consist of several components because it is hard to believe that $V$ (typically $V > 40$) variables can be represented by a single variable, even using the input-dependent attention weights. If this is the case, we should be able to split the $D$-dimensional space into $R$ spaces, where $R$ is the potential number of representative variables. As ViT has many square matrices of the size of the embedding dimension $D$, this split directly reduces the model size from $\mathcal{O}(R\times D^2/R^2)$.

We thus propose to use more than one representative variable. Hence, the embedded input $\widetilde{X} \in\mathbb{R}^{V\times D}$ is split to
\begin{align}
    \widetilde{X} = \left[\widetilde{X}_1\ \cdots \ \widetilde{X}_R \right],
\end{align}
where $\widetilde{X}_k \in \mathbb{R}^{V \times (D/R)}$ for $k=1,\ldots, R$.
We prepare trainable query vectors $\bm{q}_1, \ldots, \bm{q}_R$ and apply ClimaX variable aggregation for each $(\bm{q}_k, \widetilde{X}_k)$ pair to obtain $\bm{z}_k$. 

The embedding vectors $\{\bm{z}_k\}_k$ are then repeatedly and alternating processed by two types of transformer blocks. The first transformer block is the standard encoder layer, which consists of self-attention and feed-forward networks to extract cross-tokens and token-wise features. The second transformer block is also the standard encoder layer but introduced to allow the $R$ representative variables to interact. 
Specifically, the concatenation of $R$ representative variables is input $Z=[z_1 \cdots z_R]\in\mathbb{R}^{R \times D/R}$ to small encoder layer. The self-attention layer computes an attention map of size $R$ and mixes the representative variables.

\subsection{Regional split training}
Data-driven models have problems with high computational costs during training, which can be attributed to spatial resolution. This is because higher spatial resolution increases the number of tokens, which affects the memory cost of attention computation quadratically. However, weather forecasting at a particular point should be mainly affected the local region. Thus, we can naturally expect that training on regional input (i.e., spatially cropped input), if only the cropped regions cover the entire space as a whole, leads to a descent training, if not as successful as the global training.
Here, we examine how much time and memory reduction can be obtained from this simple strategy and how much it affects forecasting performance. In the training, an input $X_t\in\mathbb{R}^{H\times W\times V}$ is cropped into sub-region $\mathcal{C}(X_t)\in\mathbb{R}^{(H/S)\times (W/S)\times V}$, and the loss of model's output is only measured on this region. The region to crop can be randomly determined or canonically selected without overlap. This makes the cost of attention computation $\mathcal{O}(1/S^2)$-time smaller.

\begin{table*}[t]
    \centering
    \caption{Comparison of global training and regional split training by VarteX with $R=2$ and lead time $\Delta t =6$. The result of ClimaX with Global Training is provided as a reference.}
    \label{tab:comparison_training_method}
    \vskip 0.15in
    \begin{tabular}{lcccccccccc}
        \toprule
        \multirow{2}{*}{Crop size} &
        \multicolumn{2}{c}{U10} &
        \multicolumn{2}{c}{T2m} &
        \multicolumn{2}{c}{Z500} &
        \multicolumn{2}{c}{T850} &
        \multirow{1}{*}{Train time} &
        \multirow{1}{*}{Memory} \\
        & ACC $\uparrow$ & RMSE $\downarrow$& ACC $\uparrow$ & RMSE $\downarrow$& ACC $\uparrow$ & RMSE $\downarrow$& ACC $\uparrow$ & RMSE $\downarrow$ & (h) & (GB) \\ \midrule
        Global & \textbf{0.89} & \textbf{1.89}  & \textbf{0.79} & \textbf{3.24} & \textbf{0.97} & \textbf{247.33} & \textbf{0.92} & \textbf{1.90} & 14.60 & 33.02  \\
        16$\times$ 32 & 0.88 & 1.97& 0.78 & 3.39 & 0.96 & 291.19 & 0.91 & 2.08  & 6.10 & 7.70 \\ 
        8$\times$ 16 & 0.76 & 2.74& 0.63 & 5.23 &0.80 & 628.71 & 0.77 & 3.34  & 4.18 & 6.23 \\
        4$\times$8 & 0.81 & 2.48 & 0.68 & 4.57 & 0.90 & 458.73 & 0.82  & 2.96 & 4.01 & 6.11 \\ \midrule
        ClimaX & 0.59 & 3.35 & 0.66 & 4.19 & 0.76 & 667.34 & 0.71 & 3.53 & 12.28  & 21.73 \\ 
        \bottomrule
    \end{tabular}
    \vskip -0.1in
\end{table*}

\section{Experiments}\label{sec:experiments}
We now evaluate the forecasting ability and the efficiency of the proposed model, VarteX, and ClimaX as a baseline.

\subsection{Dataset and training setup}
We train VarteX and ClimaX on ERA5~\cite{hersbach2020era5} from scratch following the training setup given in~\cite{nguyen2023climax}. 

\paragraph{Dataset.} ERA5 is a publicly accessible atmospheric reanalysis dataset provided by the ECMWF. The full spatial resolution is 0.25\textdegree~($721\times 1440$ grids). As in ~\cite{nguyen2023climax}, we use ERA5 data downsampled to a spatial resolution of 5.625\textdegree~($32\times 64$ grids) provided by WeatherBench~\cite{rasp2020weatherbench}; 48 meteorological variables are used in training, and four are the target of forecast, i.e., geopotential at 500 hPa~(Z500), temperature at 850 hPa (T850), temperature at 2 meters from the ground (T2m), and zonal wind speed at 10 meters from the ground (U10). Each channel is standardized to have a mean of 0 and a standard deviation of 1. The dataset spans hourly data from 2006 to 2018, with 2006 to 2015 used for training, 2016 for validation, and 2017 to 2018 for testing. This division allows for comprehensive training and robust validation and testing of the predictive capabilities of the models involved.

\paragraph{Training.} Both VarteX and ClimaX are trained with latitude-weighted mean squared error~(MSE) loss to predict from meteorological sample $X_t \in \mathbb{R}^{H\times W \times V}$ at time $t$ to that after $\Delta t$ steps. We trained two models for lead time $\Delta t = 6$ and $24$ hours. The embedding dimension of Climax is $D = 1024$, and that of VarteX is $D/R$. Other architecture parameters, such as the number of attention heads, are all common between VarteX and ClimaX. 
Other detailed experimental settings follow those of ClimaX (cf. \cref{Training_details}).

\subsection{Effect of representative variables}\label{subsec:effect_rep_var}
\cref{Tab:Comparison_climax_vartex} compares the predictive performance of VarteX and ClimaX with 6-hour and 24-hour lead times, using RMSE and ACC metrics. 
VarteX, with two representative variables, significantly outperforms ClimaX regarding RMSE and ACC for all target variables with approximately a 45\% reduction in the model size. 
This justifies our hypothesis that explicit handling of multiple representative variables, rather than implicitly having them in a single representative variable, improves learning efficiency.
Increasing the number of representative variables in VarteX further reduces the model size; however, a drastic performance drop is observed. 
We consider that the embedding dimension $D/R$ per representative variable limits the network capacity. To examine this, we tested the case of $R=4$ again by increasing the embedding dimension (per representative variable) from $D/R$ to $2D/R$. This makes the embedding dimension the same as that in the $R=2$ case. We observed a sharp performance improvement, but this does not outperform the case of $R=2$ in both prediction ability.

\subsection{Effect of Crop Size on Predictive Performance}
We next compare standard training (referred to as global training) and regional split training. Given the results in Section~\ref{subsec:effect_rep_var}, we focus on VarteX with $R=2$.   During the training of VarteX with regional split training with $S = 2, 4, 8$, equivalent to $16\times 32$, $8\times 16$, and $4\times 8$ grids per each cropped region. At the training, the loss is computed for each grid independently, and at the inference, an input is split and fed to the model and then reconstructed from the output. Note that the region split is done canonically; if $S=2$, the input is spacially divided into top left, top right, bottom left, and bottom right. We also tested a random selection of cropping regions. However, this was not as successful as the canonical division.

\cref{tab:comparison_training_method} compares VarteX with two representative variables using regional split training and global training for a 6-hour lead time. The results indicate that a crop size of $16\times 32$ achieves the best trade-off between the performance and training cost.\footnote{Note that the optimal number of splits can depend on the resolution.} The smaller crop size deteriorates the forecasting performance with limited improvement in the training time and memory consumption. This is because the reduction in the number of tokens from $N$ to $N/2$, which quadratically reduces the size of the attention map, is large in an absolute sense, but from $N/2$ to $N/4$ is rather marginal, and other factors become a bottleneck.
To summarize, our experiments suggest that VarteX with $R=2$ and regional split training with $S=2$ (equivalent to $16 \times 32$) is the current best practice. Compared to the original ClimaX with global training, we significantly improved the ACC and RMSE with 55\% model size, 40\% training hours, and 25\% memory consumption.

\section{Conclusions}
In this study, we addressed efficient learning over many diverse meteorological variables using the ViT-based model. Inspired by ClimaX, we propose a new variable aggregation scheme explicitly modeled to extract several representative variables, significantly improving the forecasting performance and reducing the model size. Further, we examined region-wise training, which reduces training time and memory cost by a large margin at a subtle cost in the forecasting scores. While this paper focuses on training ERA5 from scratch, we may apply our results to build a foundation model by following the large-scale training of ClimaX. 

\section{Acknowledgments}
This work was supported by JST Moonshot R\&D Program Grant Number JPMJMS2389.

\bibliographystyle{icml2024}

\newpage
\appendix
\onecolumn
\section{Experiment details}
\subsection{Training details}\label{Training_details}
In this study, the experimental conditions were set up according to the paper under ClimaX. The two models, VarteX and ClimaX, are trained using an AdamW optimizer with a learning rate of $5 \times 10^{-7}$ and a weight decay of $1\times 10^{-5}$. The training schedule includes a linear warmup over five epochs followed by 45 epochs of cosine annealing. The total training period spans 50 epochs with a batch size of 128, utilizing a gradient accumulation strategy of 4 steps for every 32 batch sizes.

\subsection{Software and Hardware}
We use the ClimaX repository from GitHub (\url{https://github.com/microsoft/ClimaX}). All experiments use an NVIDIA RTX A6000 with 48GB of memory.

\subsection{Hyperparameters}
\cref{Tab:hyperparameters_of_vartex} and \cref{Tab:hyperparameters_of_climaX} present the hyperparameters of the models utilized in this experiment. The hyperparameters for ClimaX are adapted from the original paper, while VarteX shares the same values for common parameters with ClimaX. Additionally, \cref{Tab:weatherbench_variables} illustrates the meteorological variables contained within the input data.

\begin{table}[h]
\caption{Hyperparameters of VarteX.}
\label{Tab:hyperparameters_of_vartex}
\vskip 0.15in
\begin{center}
\begin{small}
\begin{tabular}{ll}
\toprule
Hyperparameter & Value  \\
\midrule
Default variables & All variables in \cref{Tab:weatherbench_variables} \\
Image size & [32, 64] \\
Patch size  & 2  \\
Embedding dimension & 1024  \\
Number of ViT blocks & 8 \\
Number of attention heads & 16 \\
Number of representative variables & 2 \\
MLP ratio & 4 \\
Prediction depth & 2 \\
Hidden dimension in prediction head & 1024 \\
Drop path & 0.1 \\
Drop rate & 0.1 \\
\bottomrule
\end{tabular}
\end{small}
\end{center}
\vskip -0.1in
\end{table}

\begin{table}[h]
\caption{Hyperparameters of ClimaX \cite{nguyen2023climax}.}
\label{Tab:hyperparameters_of_climaX}
\vskip 0.15in
\begin{center}
\begin{small}
\begin{tabular}{ll}
\toprule
Hyperparameter & Value  \\
\midrule
Default variables & All variables in \cref{Tab:weatherbench_variables} \\
Image size & [32, 64] \\
Patch size  & 2  \\
Embedding dimension & 1024  \\
Number of ViT blocks & 8 \\
Number of attention heads & 16 \\
MLP ratio & 4 \\
Prediction depth & 2 \\
Hidden dimension in prediction head & 1024 \\
Drop path & 0.1 \\
Drop rate & 0.1 \\
\bottomrule
\end{tabular}
\end{small}
\end{center}
\vskip -0.1in
\end{table}

\begin{table}[H]
\caption{The variables from WeatherBench used in the model. The same variables are utilized as in ClimaX.}
\label{Tab:weatherbench_variables}
\vskip 0.15in
\begin{center}
\begin{small}
\begin{tabular}{lll}
\toprule
Variable name & Abbrev. & Pressure levels  \\
\midrule
Land-sea mask & LSM & \\
Orography & & \\
2 metre temperature & T2m &\\
10 meter U wind component & U10 &\\
10 meter V wind component & V10 &\\
\midrule
Geopotential & Z & 50, 250, 500, 600, 700, 850, 925 \\
U wind component & U & 50, 250, 500, 600, 700, 850, 925 \\
V wind component & V & 50, 250, 500, 600, 700, 850, 925 \\
Temperature & T & 50, 250, 500, 600, 700, 850, 925 \\
Specific humidity & Q & 50, 250, 500, 600, 700, 850, 925 \\
Relative humidity & R & 50, 250, 500, 600, 700, 850, 925 \\
\bottomrule
\end{tabular}
\end{small}
\end{center}
\vskip -0.1in
\end{table}

\subsection{Loss function and Metrics}
This section presents the evaluation metrics used in the experiment. $\widetilde{Y}$ and $Y$ represent the forecast and ground truth, respectively, while K denotes the total number of test data points. Additionally, C represents climatology, defined as the time average over the entire test data set, $C=1/K\sum_{k}Y_{k}$.

\textbf{Latitude weighting factor}
\begin{equation}
  L(h)=\frac{\cos{(h)}}{\frac{1}{H}\sum_{lat=1}^{H}\cos{(lat)}}
\end{equation}

\textbf{Latitude-weighted mean square error (MSE)}
\begin{equation}
  \text{MSE}=\mathbb{E}\Big\lbrack\frac{1}{H\times W}\sum_{h=1}^{H}\sum_{w=1}^{W}L(h)(\widetilde{Y}_{k,h,w}-Y_{k,h,w})^2\Big\rbrack.
\end{equation}

\textbf{Latitude-weighted root mean square error (RMSE)}
\begin{equation}
  \text{RMSE}=\frac{1}{K}\sum_{k=1}^{K}\sqrt{\frac{1}{H\times W}\sum_{h=1}^{H}\sum_{w=1}^{W}L(h)(\widetilde{Y}_{k,h,w}-Y_{k,h,w})^2}.
\end{equation}

\textbf{Latitude-weighted anomaly correlation coefficient (ACC)}
\begin{equation}
  \text{ACC}=\frac{\sum_{k,h,w}L(h)\widetilde{Y}'_{k,h,w}Y'_{k,h,w}}{\sqrt{\sum_{k,h,w}L(h)\widetilde{Y}'^{2}_{k,h,w}\sum_{k,h,w}L(h)Y'^{2}_{k,h,w}}}
\end{equation}
\begin{equation}
 \widetilde{Y}' = \widetilde{Y} - C, Y'=Y-C
\end{equation}

\section{Qualitative evaluation}
The qualitative evaluation of the predictive performance of VarteX for all target variables with a 6-hour lead time is shown. The results include VarteX with both global training and regional split training. The first column shows the initial state, the second column presents the ground truth, the third column displays the predicted results, and the fourth column indicates the bias between the ground truth and the predictions. Note that these visualizations are for reference only and that this experiment was not pre-trained in the same way as ClimaX. We believe that better visualization can be obtained if pre-training is used as in the ClimaX paper~\cite{nguyen2023climax}.

\begin{figure}[t]
\vskip 0.2in
\begin{center}
\centerline{\includegraphics[width=\columnwidth]{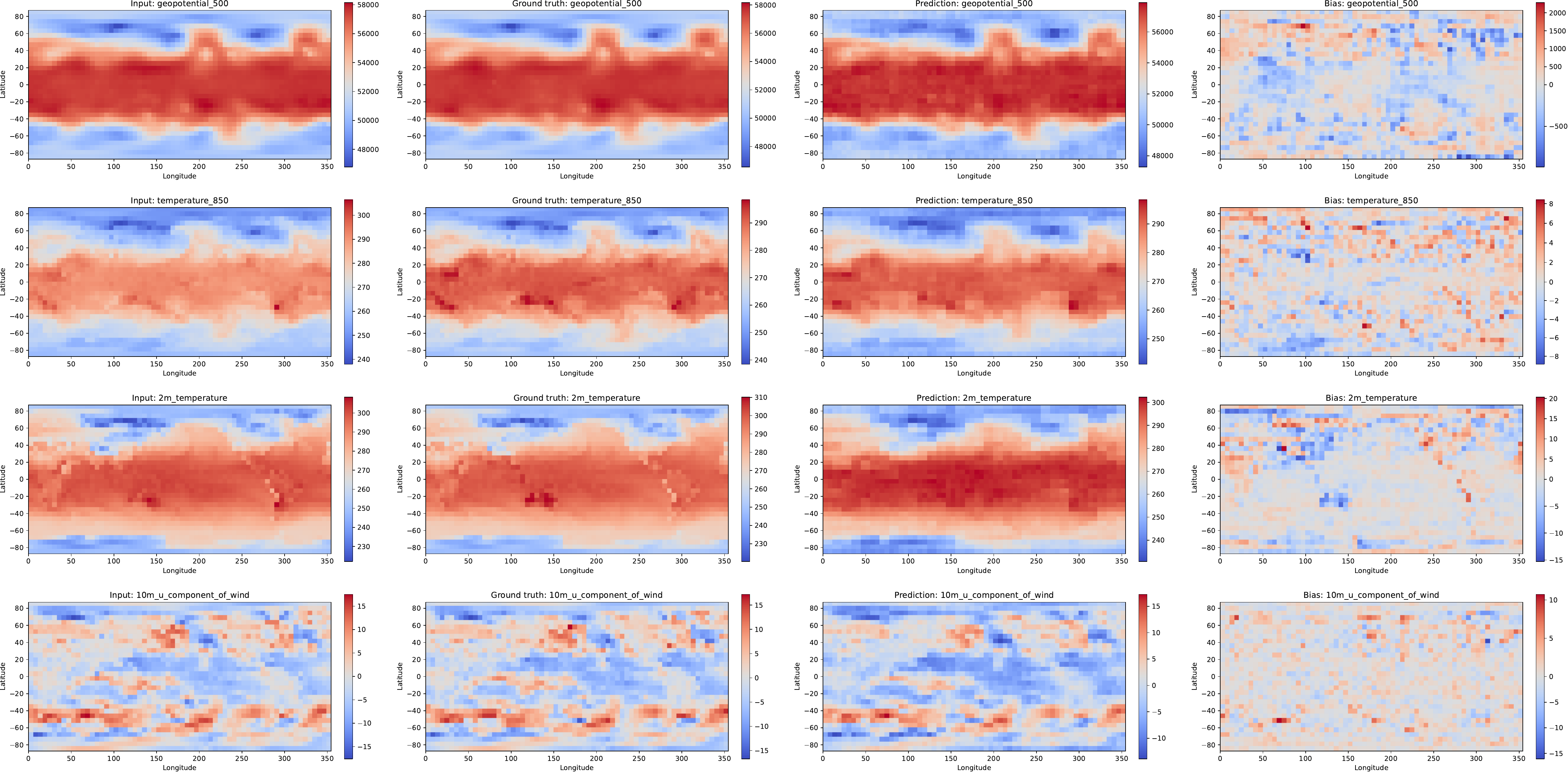}}
\caption{An example of VarteX's forecasting results with two representative variables and the Ground Truth for a 6-hour lead time.}
\label{fig:vartex_r2_qualitative}
\end{center}
\vskip -0.2in
\end{figure}

\begin{figure}[t]
\vskip 0.2in
\begin{center}
\centerline{\includegraphics[width=\columnwidth]{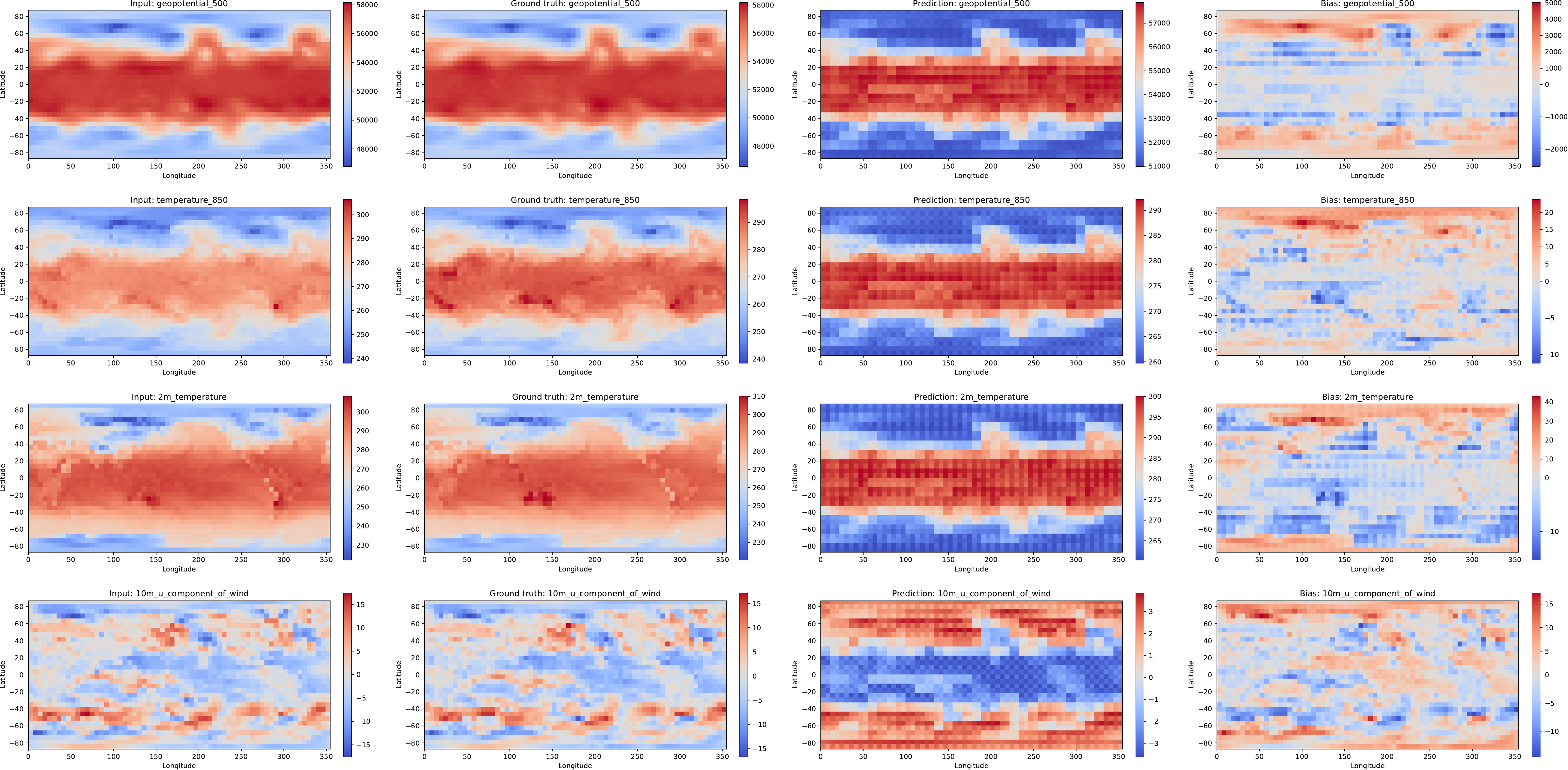}}
\caption{An example of VarteX's forecasting results with four representative variables and the Ground Truth for a 6-hour lead time.}
\label{fig:vartex_r4_qualitative}
\end{center}
\vskip -0.2in
\end{figure}

\begin{figure}[t]
\vskip 0.2in
\begin{center}
\centerline{\includegraphics[width=\columnwidth]{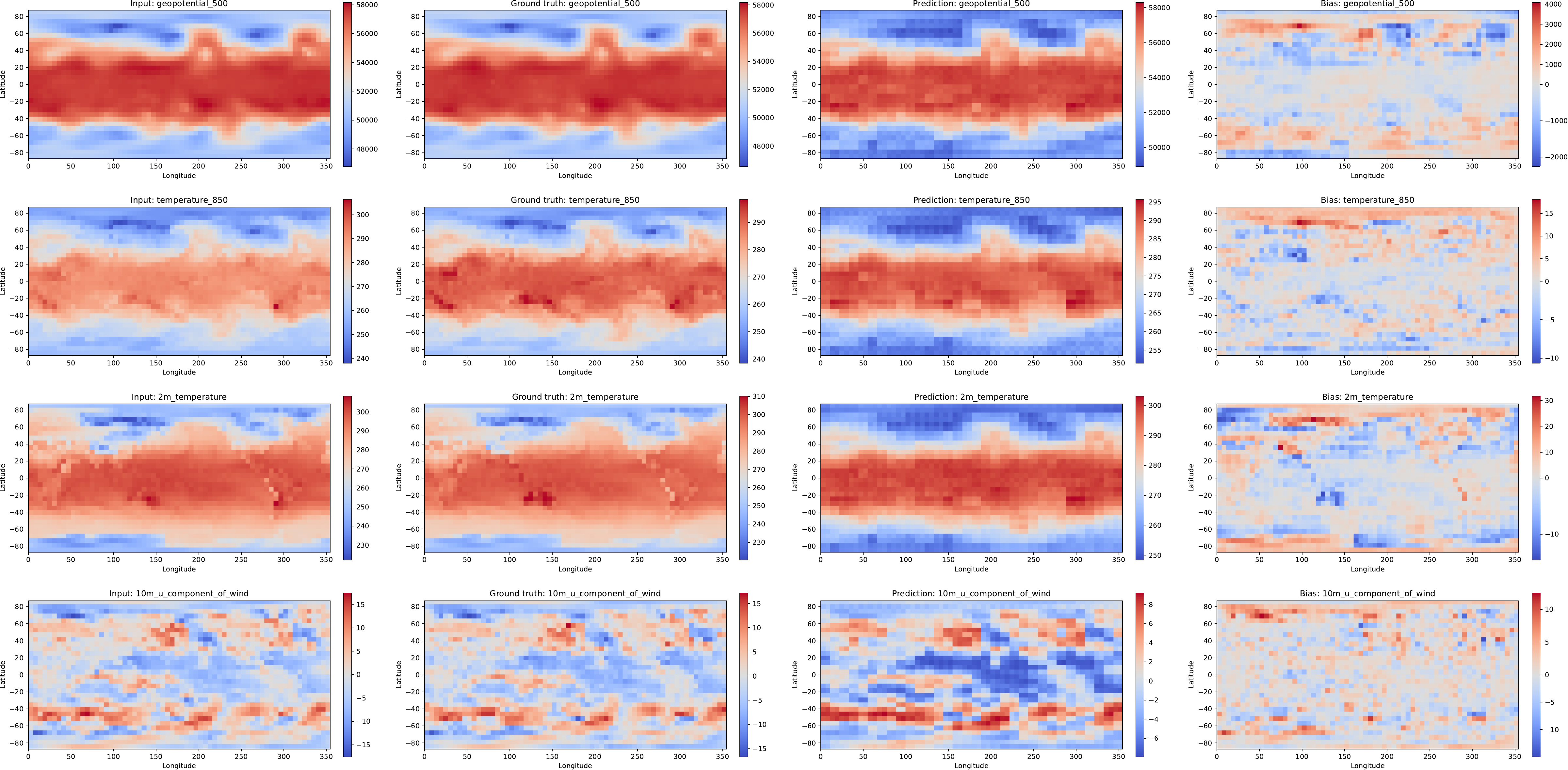}}
\caption{
An example of VarteX's forecasting results with two representative variables and the Ground Truth for a 6-hour lead time, with the embedding dimension specifically set to 2048.}
\label{fig:vartex_r4_d_2048_qualitative}
\end{center}
\vskip -0.2in
\end{figure}

\begin{figure}[t]
\vskip 0.2in
\begin{center}
\centerline{\includegraphics[width=\columnwidth]{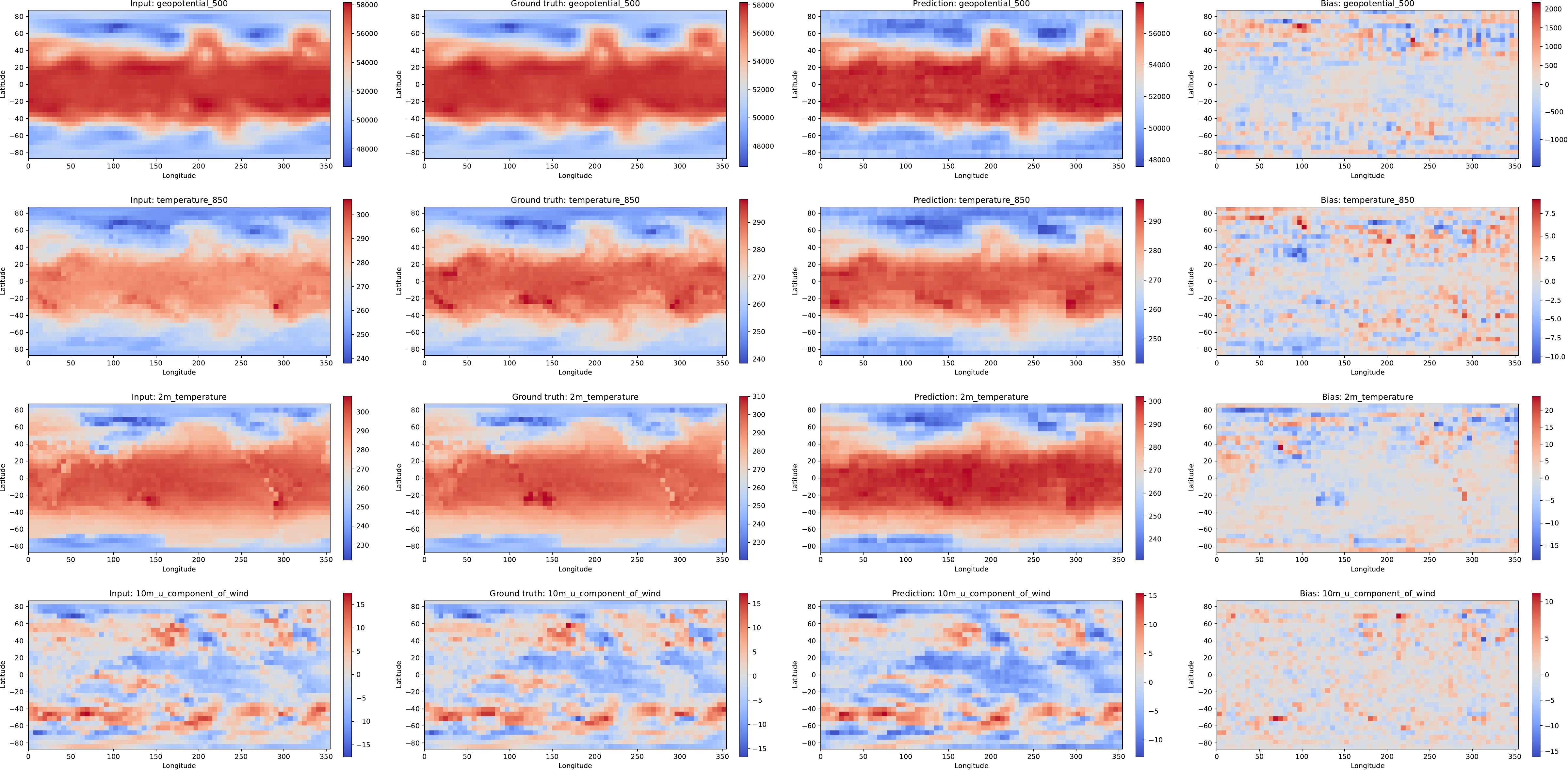}}
\caption{
An example of VarteX's forecasting results and Ground Truth for a 6-hour lead time using a $16\times 32$ crop size with regional split training.}
\label{fig:split_16_32_vartex_r2_qualitative}
\end{center}
\vskip -0.2in
\end{figure}

\begin{figure}[t]
\vskip 0.2in
\begin{center}
\centerline{\includegraphics[width=\columnwidth]{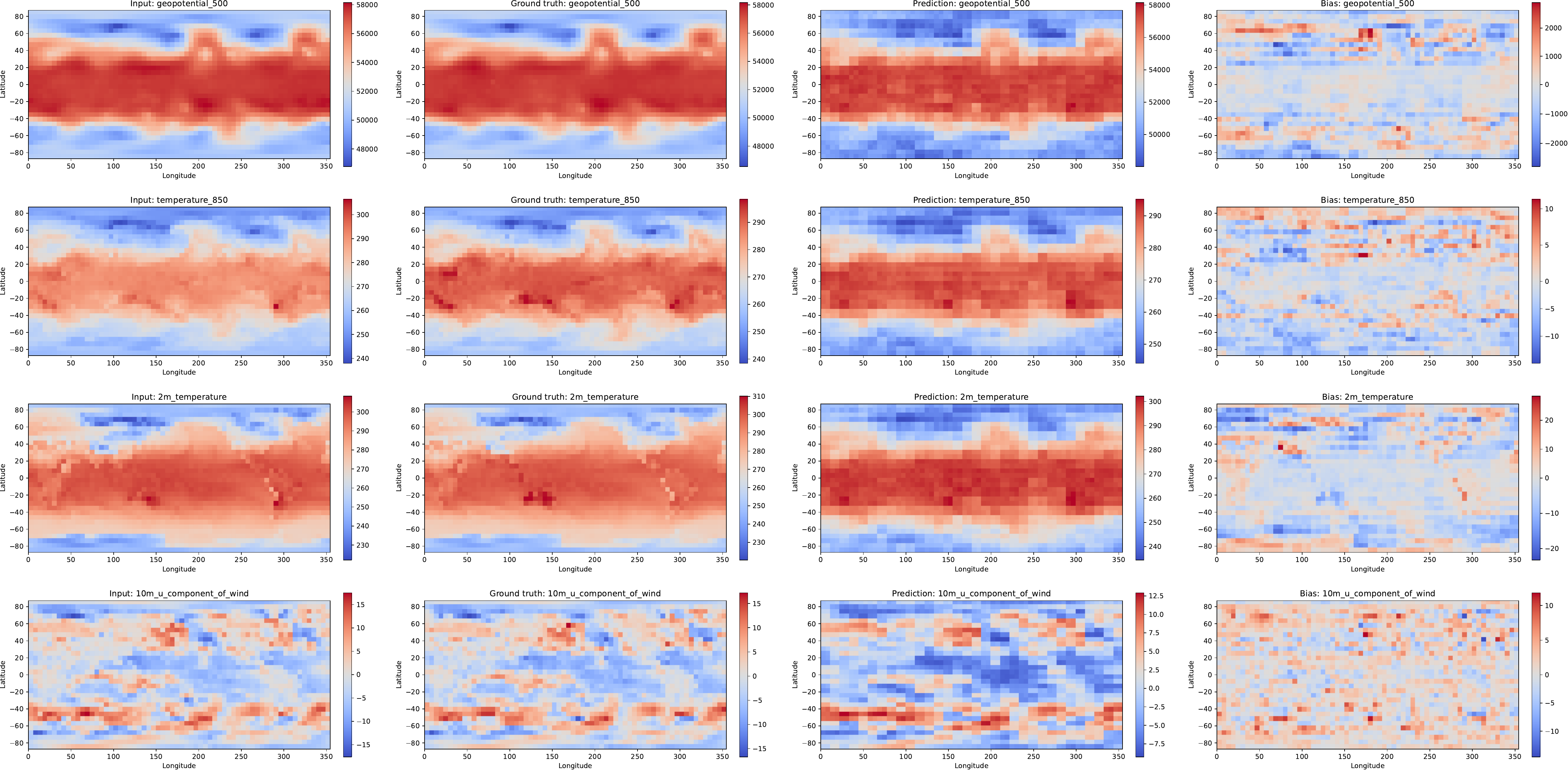}}
\caption{
An example of VarteX's forecasting results and Ground Truth for a 6-hour lead time using a $8\times 16$ crop size with regional split training.}
\label{fig:split_8_16_vartex_r2_qualitative}
\end{center}
\vskip -0.2in
\end{figure}

\begin{figure}[t]
\vskip 0.2in
\begin{center}
\centerline{\includegraphics[width=\columnwidth]{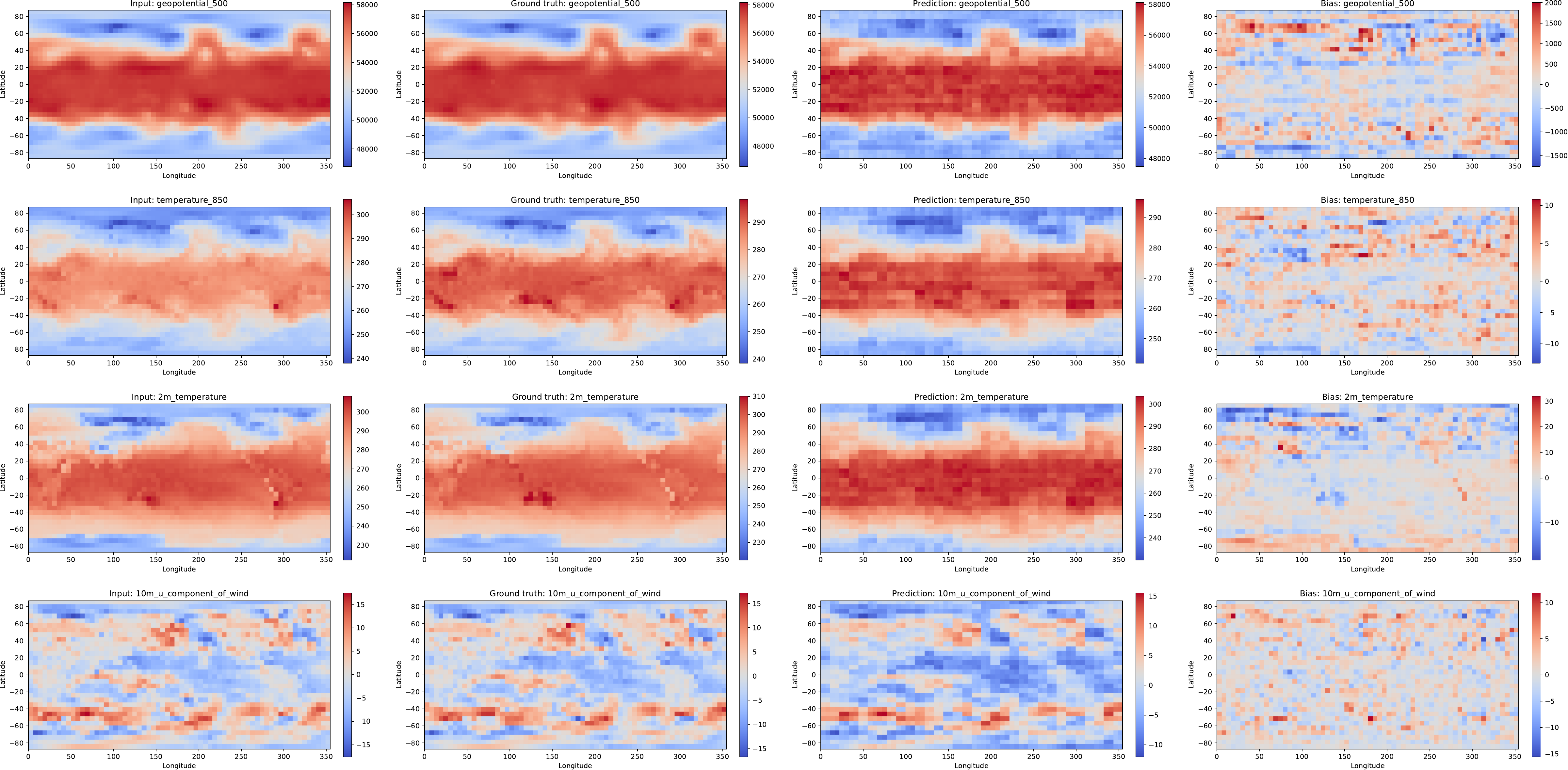}}
\caption{
An example of VarteX's forecasting results and Ground Truth for a 6-hour lead time using a $4\times 8$ crop size with regional split training.}
\label{fig:split_4_8_vartex_r2_qualitative}
\end{center}
\vskip -0.2in
\end{figure}

\end{document}